\documentclass[11pt]{article}
\usepackage{acl}
\usepackage{times}
\usepackage{latexsym}
\usepackage[T1]{fontenc}
\usepackage[utf8]{inputenc}
\usepackage{microtype}
\usepackage{inconsolata}

\usepackage{booktabs}
\usepackage{algorithm}
\usepackage{algpseudocode}
\usepackage{amsmath}
\usepackage{amssymb}
\usepackage{mathtools}
\usepackage{mathrsfs}
\usepackage{amsthm}
\usepackage{graphicx}
\usepackage{subcaption}
\usepackage{tikz}
\usepackage{pgfplots}
\usepackage{hyperref}
\pgfplotsset{compat=1.18}

\definecolor{morandiBlue}{HTML}{6D8FA3}
\definecolor{morandiGray}{HTML}{B8B0A3}
\definecolor{morandiSage}{HTML}{8FA58E}
\definecolor{morandiOchre}{HTML}{C7B37F}
\definecolor{morandiRose}{HTML}{C9A7A0}
\definecolor{morandiPurple}{HTML}{A89CB3}
\definecolor{morandiBlue}{HTML}{6D8FA3}
\definecolor{morandiRose}{HTML}{C9A7A0}
\usepackage{tcolorbox}
\usepackage{tabularx}
\usepackage{multirow}
\usepackage{siunitx}
\usepackage{adjustbox}          

\usetikzlibrary{shapes,arrows,positioning,decorations.pathmorphing}

\mathtoolsset{showonlyrefs}

\newcommand{\method}{AgentMob}

\definecolor{HeaderBlue}{HTML}{0E5E76}
\definecolor{HeaderGray}{HTML}{D9DDE0}
\definecolor{HeaderLight}{HTML}{ECEFF1}

\title{Towards Efficient and Evidence-grounded Mobility Prediction \\
with LLM-Driven Agents}

\author{
  Linyao Chen\textsuperscript{1} \quad
  Qinlao Zhao\textsuperscript{2} \quad
  Zechen Li\textsuperscript{3} \quad
  Mingming Li\textsuperscript{1} \\
  \textbf{Likun Ni\textsuperscript{5}} \quad
  \textbf{Jinyu Chen\textsuperscript{1,$\dag$}} \quad
  \textbf{Yuhao Yao\textsuperscript{4,$\dag$}} \\
  \textbf{Xuan Song\textsuperscript{7}} \quad
  \textbf{Noboru Koshizuka\textsuperscript{1}} \quad
  \textbf{Hiroki H.\ Kobayashi\textsuperscript{1}} \\[1pt]
  \textsuperscript{1}The University of Tokyo \enspace
  \textsuperscript{2}Huazhong University of Science and Technology \\[-1pt]
  \textsuperscript{3}University of New South Wales, Sydney \enspace
  \textsuperscript{4}LocationMind Inc. \\[-1pt]
  \textsuperscript{5}Southern University of Science and Technology \\[-1pt]
  \textsuperscript{7}Jilin University \enspace
  \textsuperscript{$\dag$}Corresponding authors
}

\begin{document}
\maketitle


\begin{abstract}
Individual-level mobility prediction is central to urban simulation, transportation planning, and policy analysis. Supervised sequence models achieve strong accuracy but require task-specific training and offer limited decision-level transparency. Recent LLM-based methods improve interpretability, yet mostly rely on static prompts and single-pass inference, limiting their ability to seek additional evidence when mobility signals are weak or conflicting. We propose \method{}, a training-free LLM-driven agent framework that formulates next-location prediction as adaptive evidence-controlled decision making. \method{} resolves routine cases through a fast path based on historical regularity, while ambiguous cases trigger iterative tool use over recent trajectories, historical behavior, stay-move likelihood, and geographical evidence. Across three mobility datasets, \method{} achieves the strongest overall performance among training-free LLM-based methods, with GPT-5.4 reaching 71.42\% Acc@1 on BW, 33.14\% on YJMob100K, and 33.50\% on Shanghai ISP. On BW non-fast-path cases, the LLM controller improves Acc@1 from 30.65\% to 48.62\% over a same-tool statistical baseline, showing that its main benefit lies in resolving ambiguous predictions through adaptive evidence gathering. Our code is available at \href{https://github.com/Unknown-zoo/AgentMob}{AgentMob}.
\end{abstract}

  \begin{figure}[t]
    \centering
    \includegraphics[width=\columnwidth]{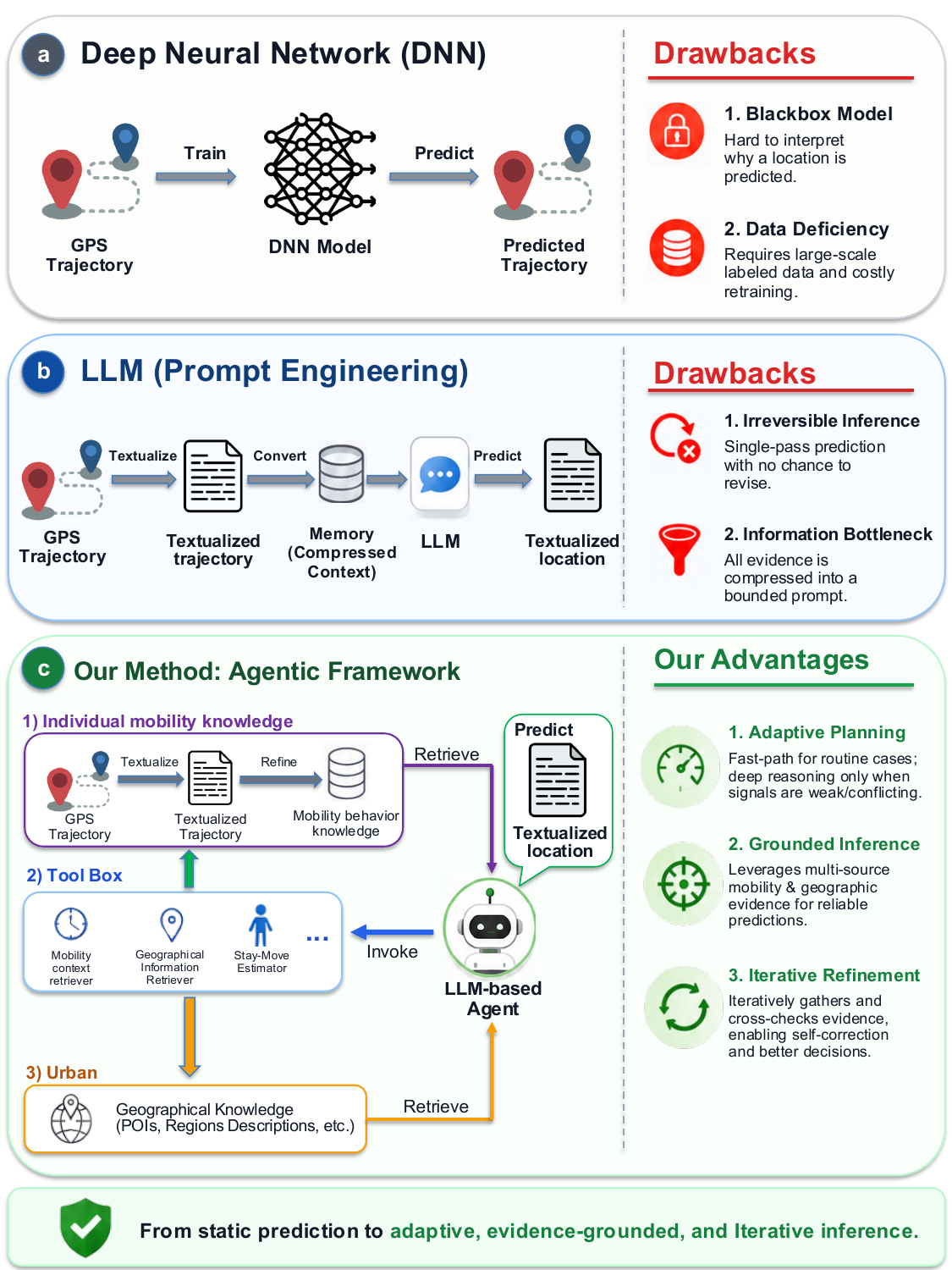}
    \caption{Comparison between different mobility prediction paradigms. Our method proposes adaptive, evidance-grounded and iterative prediction in a training-free manner, provides effecient and reliable predictions.}
    \vspace{-1em}
    \label{fig:teaser}
\end{figure}  
\section{Introduction}\label{sec:intro}


Accurate human mobility prediction is fundamental to urban planning~\citep{zheng2014urban}, transportation management~\citep{luca2021survey}, and public health analysis~\citep{barbosa2020scales}. At the individual level, next-location prediction supports location-based services~\citep{zheng2015trajectory}, demand-aware transportation systems, and targeted public health interventions~\citep{oliver2020mobile}. The core challenge is to infer a user's next spatial state from historical trajectories, recent movement context, and the geographical semantics of candidate locations.

Existing approaches largely follow two paradigms. Supervised sequence models, including RNNs~\citep{feng2018deepmove}, Transformers~\citep{vaswani2017attention}, and their variants, learn mobility regularities from large-scale trajectory data and can achieve strong predictive accuracy. However, they require task-specific training, are costly to adapt to new cities or spatial granularities, and typically provide limited insight into why a particular location is predicted. Recent LLM-based methods~\citep{zhao2026llm,zhang2024tinyllama} offer a more interpretable alternative by converting trajectories into text and leveraging language-model reasoning. Fine-tuning approaches~\citep{feng2024limp,llm4poi} still inherit substantial training cost, while prompt-based methods~\citep{wang2024llmmob,calderon2025cognitive} avoid retraining but compress all available evidence into a static prompt. Agent-like mobility frameworks~\citep{feng2025agentmove,li2025hierarchicalllmagent} introduce structured memory or reasoning modules, but often rely on pre-scripted workflows with limited ability to adapt the amount and type of evidence used for each prediction.


This static prediction paradigm is problematic since mobility instances are not equally difficult. Many cases are routine: a user repeatedly visits the same location at a similar weekday and hour, so historical regularity may be sufficient. Other cases are ambiguous: recent movement, long-term routines, stay--move tendencies, and geographical plausibility may point to different candidate locations. In such cases, a reliable predictor should not commit after a single forward pass. It should retrieve targeted evidence, cross-check conflicting signals, and stop once the evidence is sufficient. Existing prompt-only LLM methods lack this capability since prediction is performed through a fixed context and a single generation step, with no structured mechanism to verify or revise the decision.



To address this limitation, we propose \method{}, a training-free agentic framework that formulates next-location prediction as \emph{adaptive evidence-controlled decision making}. Instead of treating the LLM as a direct trajectory predictor, \method{} uses it as a controller that decides how much evidence is needed for each instance. Routine cases exit through a fast path based on strong historical regularity, while ambiguous cases trigger iterative tool use over recent trajectory context, historical behavioral statistics, stay--move likelihood, geographical distance, and location semantics. The final prediction is thus grounded in explicit tool outputs rather than opaque prompt-only generation.

As illustrated in Figure~\ref{fig:teaser}, \method{} differs from supervised DNNs and prompt-based LLM methods in both computation and reasoning. It does not require task-specific model training, and it avoids forcing all samples through the same fixed inference procedure. Instead, it allocates reasoning effort according to prediction difficulty and records timestamp-bounded tool outputs for auditability. We evaluate \method{} with both open-source and closed-source LLM backbones on three mobility datasets with different spatial granularities. Experiments show that \method{} achieves the strongest overall performance among training-free LLM-based methods, while additional analyses demonstrate that the LLM controller is most beneficial when deterministic mobility statistics are insufficient or conflicting. In summary, our contributions are as follows:

\begin{itemize}
    \item \textbf{Adaptive evidence-control formulation.}
    We recast next-location prediction as an instance-level decision process, where an LLM agent chooses between a fast historical-regularity path and additional evidence gathering for ambiguous cases.

    \item \textbf{Training-free tool-augmented mobility agent.}
    We develop \method{}, which selectively invokes mobility tools over recent context, historical behavior, stay--move likelihood, and geographical evidence to produce auditable prediction traces.

    \item \textbf{Strong and explainable training-free performance.}
    Across three mobility datasets and multiple LLM backbones, \method{} achieves the strongest overall results among training-free LLM-based methods, with analyses showing that adaptive evidence gathering is most useful when mobility signals conflict.
\end{itemize}

\section{Related Work}\label{sec:related}

\paragraph{Individual Next-Location Prediction.}
Next-location prediction has evolved from Markov-chain and factorization-based methods~\citep{rendle2010factorizing,markov} to deep sequence models based on RNNs~\citep{feng2018deepmove,rnn2}, Transformers~\citep{lian2020geography,luo2021stan,qin2022next,sun2024going}, and GNNs~\citep{wang2024spatiotemporal,wu2024imitate}. These methods learn complex spatiotemporal dependencies from large-scale mobility data, but typically require task-specific training and provide limited decision-level transparency. Recent LLM-based methods convert trajectories into textual inputs for mobility prediction or generation~\citep{wang2023would,wang2024large,liang2024exploring}; some fine-tune LLMs for point-of-interest recommendation or trajectory generation~\citep{llm4poi,geollama}, while others integrate memory, urban knowledge, or structured reasoning modules~\citep{feng2025agentmove,ju2025trajllm,zhong2025comapoi,liu2025gatsim}. However, most existing LLM-based mobility methods still follow a fixed inference procedure, where history or knowledge is retrieved in advance, compressed into a prompt or scripted workflow, and then used to produce a prediction. Closely related agentic mobility frameworks, such as AgentMove~\citep{feng2025agentmove} and ARMove~\citep{wang2026armove}, introduce structured memory, user profiling, feature management, or feature optimization. In contrast, \method{} treats next-location prediction as an instance-level evidence-control problem: the agent decides whether a sample can exit through a fast historical-regularity path, which tools to invoke for ambiguous cases, and how to resolve conflicts among recent context, historical behavior, stay--move likelihood, and geographical plausibility.

\paragraph{LLM-Driven Agents and Evidence-Grounded Tool Use.}
LLM-driven agents extend LMs from passive text generation to interactive reasoning and action. Prior work has explored instruction-following~\citep{ouyang2022instructgpt}, multi-agent collaboration~\citep{li2023camel,wu2023autogen,li2026zara}, general task automation~\citep{hu2025owl,tang2025agentkb}, social simulation~\citep{park2023generative}, and tool-augmented problem solving~\citep{nakano2021webgpt,qin2023toolllm}. These studies suggest that LLMs can coordinate intermediate evidence and external tools, but they are not directly designed for next-location prediction under spatiotemporal uncertainty. \method{} instantiates evidence-grounded tool use for mobility prediction, where tool invocation is controlled by prediction difficulty and each decision is traceable through timestamp-bounded behavioral and geographical evidence.
\section{Methodology}
\label{sec:method}

\subsection{Pipeline of LLM-based Tool Agent for Next Location Prediction}
\label{sec:pipeline}

\begin{figure*}[t]
    \centering
    \includegraphics[width=\textwidth]{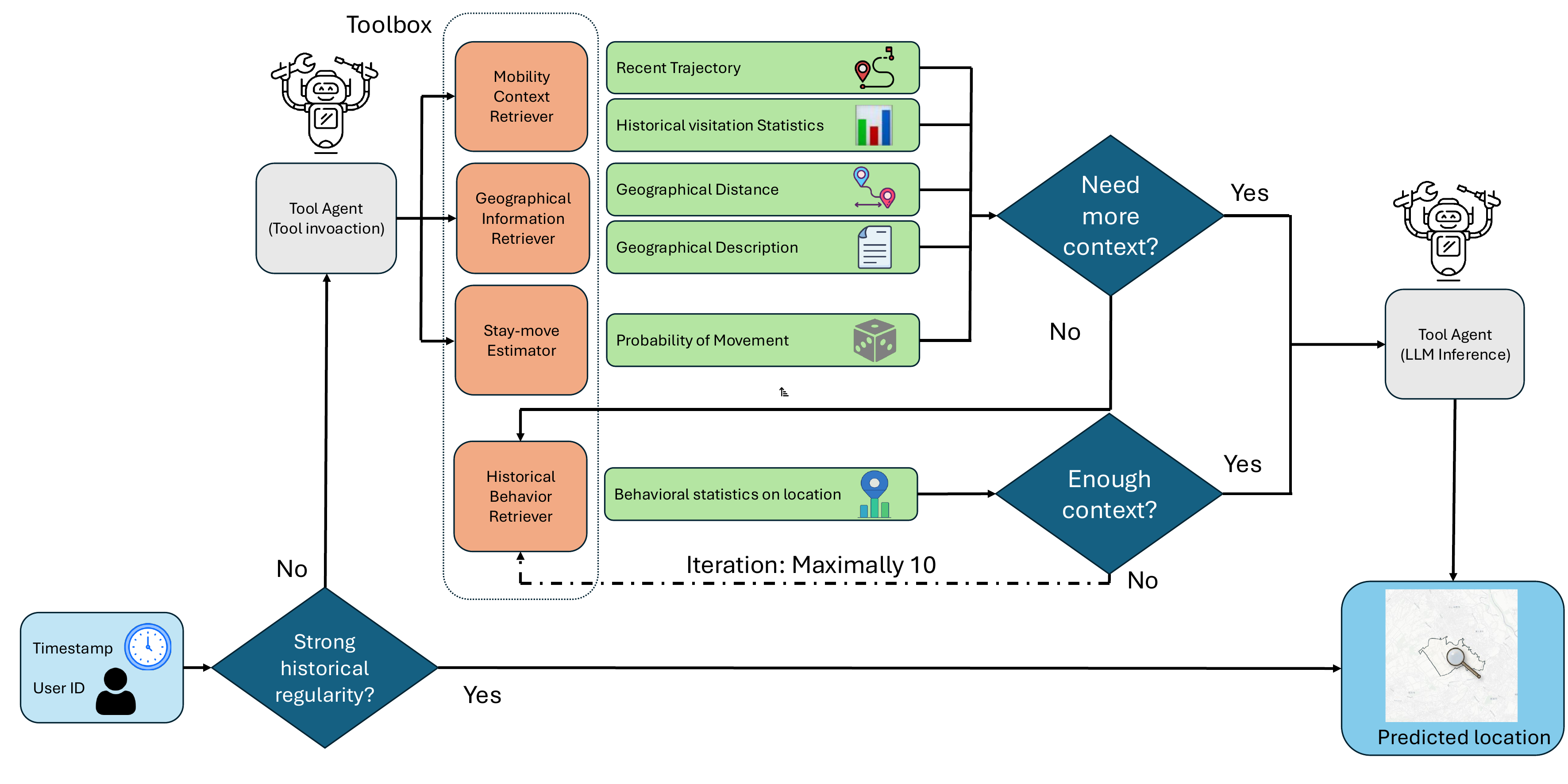}
    \caption{\method{} The workflow of Agentmob. The key faeture includes: 1. fast-path prediction for highly regular cases 2. adaptive tool use for ambiguous cases, and evidence-controlled stopping before final prediction 3. The primary tool retrieves multi-dimensional information for reliable evidence-grounded inference.}
    \label{fig:method}
    \vspace{-0.5em}
\end{figure*}


Figure~\ref{fig:method} illustrates the overall workflow of \method{}. We formulate next-location prediction as an adaptive evidence-control process, where an LLM agent decides whether a test instance can be solved from historical regularity or requires additional evidence from mobility-analysis tools. The framework is implemented with smolagents~\citep{smolagents}, but the key design is independent of a specific agent library: the LLM acts as a controller that selects tools, interprets their outputs, and produces a ranked prediction over candidate spatial units.

Given raw GPS records, we first discretize coordinates into spatial units, either administrative polygons or uniform grid cells depending on the dataset. Each spatial unit is paired with a concise textual description of its urban function, allowing the LLM to reason over both trajectory statistics and location semantics. The prediction task is then defined as selecting the spatial unit that the user will visit next, given the user ID, target timestamp, historical trajectory records, and the observed context before the target time.




\paragraph{Fast-path Prediction.}
\method{} first performs a lightweight regularity check before invoking the full tool-calling loop. If the user's historical records show a dominant location for the same weekday and hour, the agent returns this location directly as the prediction. This fast path resolves routine cases efficiently and prevents unnecessary LLM/tool computation when the historical signal is already strong.

\paragraph{Adaptive Tool-calling Prediction.}
When strong historical regularity is absent, \method{} enters the tool-calling mode. The agent invokes tools from the mobility-analysis toolbox described in Section~\ref{sec:tools}. These tools provide two complementary types of evidence: behavioral evidence, such as recent trajectory context, same-time visitation statistics, stay--move likelihood, and historical transitions under similar conditions; and geographical evidence, such as distance to candidate locations and textual descriptions of their urban functions. Based on the evidence collected so far, the agent may request additional tool outputs, compare competing candidates, or proceed to a final decision. The loop is capped at ten iterations to bound inference cost.

\paragraph{Prediction and Auditability.}
After evidence gathering, the agent outputs a final top-1 prediction together with a ranked top-$K$ candidate list for rank-based evaluation. All tool invocations are constrained to the training split and to observations available before the target timestamp, ensuring chronological validity. For auditability, each prediction trace stores the user ID, target timestamp, allowed history range, invoked tools, serialized tool outputs, LLM reasoning steps, and final ranked prediction.

\subsection{Mobility Analysis Toolbox}
\label{sec:tools}

The toolbox exposes compact, timestamp-bounded evidence that the agent can selectively query, with each tool summarizing a distinct mobility signal instead of passing raw trajectories directly to the LLM. Together, these tools cover four complementary aspects of next-location prediction: recent context, geographical plausibility, stay--move uncertainty, and historical behavior under comparable conditions.

\paragraph{Mobility Context Retriever.}
Given a target timestamp, this tool summarizes the user's short-term movement context and routine temporal patterns. It returns the recent sequence of visited locations over the past several hours, together with visitation statistics for the same weekday and hour in the user's historical records. This evidence helps the agent distinguish immediate movement continuity from long-term temporal regularity.

\paragraph{Geographical Information Retriever.}
This tool provides spatial and semantic evidence for candidate locations. It returns the distance from the current location to each candidate, as well as a concise textual description of the candidate's urban function. This allows the agent to check whether a candidate is geographically plausible and semantically consistent with the user's mobility context.

\paragraph{Stay--Move Estimator.}
This tool estimates whether the user is more likely to remain at the current location or move elsewhere. It compares the current stay duration with historical stay-duration statistics at the same location and returns a movement likelihood derived from past behavior. This evidence is especially useful when recent continuity conflicts with transition-based predictions.

\paragraph{Historical Behavior Retriever.}
This tool retrieves behavioral evidence from past visits to the same or nearby locations under comparable temporal contexts, such as similar time of day or day of week. It summarizes typical dwell times, frequent next destinations, transition tendencies, and visit regularity. The agent uses this evidence to verify uncertain candidates and resolve conflicts among competing mobility signals.

\section{Experiment}
\label{sec:exp}

\subsection{Experimental Setup}
\label{sec:setup}

\paragraph{Datasets.}
We evaluate \method{} on three mobility datasets with different spatial granularities and observation mechanisms. BW~\citep{blogwatcher2024} is a mobile-phone GPS dataset from the Tokyo metropolitan area, where raw GPS records are mapped to third-level administrative polygons annotated with hierarchical region names and functional descriptions. YJMob100K~\citep{yabe2024yjmob100k} is a large-scale mobile-phone GPS dataset discretized into anonymized 500\,m$\times$500\,m grid cells; since cells do not have real place names, we generate textual descriptions from the provided POI category distributions using LLM-based POI summarization (Appendix~\ref{app:loc_desc}). For BW and YJMob100K, we select the 100 most active users and convert continuous trajectories into spatial-unit sequences with location descriptions (Table~\ref{tab:dataset_complexity}). Shanghai ISP~\citep{feng2019dplink} is an anonymized mobile-network trajectory benchmark collected from mobile network logs in Shanghai, containing 325,215 records from April 19 to 26, 2016; following recent LLM-based next-location prediction work~\citep{feng2025agentmove}, we discretize base-station coordinates into 500\,m$\times$500\,m grid cells and generate textual descriptions for each cell.



\begin{table}[t]
\centering
\small
\resizebox{\columnwidth}{!}{
\begin{tabular}{lccc}
\toprule
\textbf{Statistic} & \textbf{BW} & \textbf{YJMob100K} & \textbf{Shanghai ISP} \\
\midrule
Records & 427,248 & 310,546 & 4,944 \\
Region & Tokyo & Anonymized & Shanghai \\
Spatial unit & Admin. polygons & 500\,m grids & 500\,m grids \\
Eval. setting & Top-100 users & Top-100 users & 200 first-test users \\
Visited locations & 4,188 polygons & 12,430 cells & 1,385 cells \\
Move / stay transitions & 28.6 / 71.4\% & 74.7 / 25.3\% & 90.0 / 10.0\% \\
Avg. unique locations & 104.8 & 413.7 & 10.7 \\
\bottomrule
\end{tabular}
}
\caption{Dataset statistics and evaluation settings.}
\label{tab:dataset_complexity}
\vspace{-1em}
\end{table}

\begin{table*}[t]

 \caption{Main performance comparison under chronological evaluation on BW, YJMob100K, and the Shanghai ISP first-test-point setting. \textbf{Bold} marks the best result among all methods, and \underline{underlining} marks the best result among training-free LLM-based methods.}
 \label{tab:professional_alignment}
 \centering
 \scriptsize
 \begin{tabular}{ll ccc ccc ccc}
 \toprule
 \multirow{2}{*}{\textbf{Method}} & \multirow{2}{*}{\textbf{Backbone}} & \multicolumn{3}{c}{\textbf{BW}} & \multicolumn{3}{c}{\textbf{YJMob100K}} & \multicolumn{3}{c}{\textbf{Shanghai ISP}} \\
 \cmidrule(lr){3-5} \cmidrule(lr){6-8} \cmidrule(l){9-11}
 & & Acc@1$\uparrow$ & MRR@5$\uparrow$ & Dist.$\downarrow$ 
  & Acc@1$\uparrow$ & MRR@5$\uparrow$ & Dist.$\downarrow$ 
  & Acc@1$\uparrow$ & MRR@5$\uparrow$ & Dist.$\downarrow$ \\
 \midrule
 \multirow{2}{*}{\textbf{Deep Neural Network}}
 & DeepMove & 72.13\% & 78.92\% & 2.21 & 31.52\% & 44.86\% & 4.86 & 20.50\% & 30.00\% & 6.95 \\
 & Transformer & \textbf{73.04}\% & \textbf{79.67}\% & \textbf{2.18} & 33.12\% & 46.12\% & 4.81 & 22.00\% & 32.25\% & 9.26 \\
\addlinespace
\multirow{3}{*}{\textbf{AgentMove}}
 & GPT-4.1-mini & 58.52\% & 70.16\% & 3.53 & 29.25\% & 41.33\% & 4.77 & 26.50\% & 39.26\% & 4.78 \\
 & Qwen3-8B & 52.58\% & 64.09\% & 4.21 & 29.41\% & 41.84\% & 5.36 & 28.00\% & 38.48\% & 5.75 \\
 & GPT-5.4 & 64.20\% & 74.02\% & 3.07 & 33.09\% & 46.51\% & 4.46 & 27.00\% & 39.28\% & 5.35 \\
\addlinespace
\multirow{3}{*}{\textbf{LLM-Mob}}
 & GPT-4.1-mini & 55.08\% & 66.10\% & 3.52 & 21.38\% & 35.42\% & 5.98 & 33.00\% & \textbf{\underline{47.61\%}} & 5.22 \\
 & Qwen3-8B & 56.19\% & 65.53\% & 3.45 & 23.57\% & 36.50\% & 5.84 & 26.50\% & 40.67\% & 6.87 \\
 & GPT-5.4 & 66.48\% & 74.84\% & 2.52 & 26.20\% & 39.15\% & 5.20 & 29.50\% & 45.38\% & 6.44 \\
\addlinespace
\multirow{3}{*}{\textbf{TrajLLM}}
 & GPT-4.1-mini & 53.08\% & 64.33\% & 6.58 & 24.25\% & 37.21\% & 11.57 & 22.50\% & 36.30\% & 7.18 \\
 & Qwen3-8B & 52.58\% & 64.04\% & 4.71 & 28.12\% & 42.23\% & 5.16 & 22.00\% & 35.00\% & 6.54 \\
 & GPT-5.4 & 62.23\% & 72.54\% & 3.48 & 29.35\% & 43.20\% & 4.68 & 25.00\% & 39.46\% & 19.74 \\
\addlinespace
\multirow{3}{*}{\textbf{LLM Urban Res.}}
 & GPT-4.1-mini & 64.38\% & 72.89\% & 2.59 & 26.59\% & 40.02\% & 5.21 & 32.00\% & 47.28\% & \textbf{\underline{4.23}} \\
 & Qwen3-8B & 58.93\% & 68.00\% & 3.29 & 25.89\% & 39.51\% & 5.90 & 19.5\% & 24.74\% & 7.37 \\
 & GPT-5.4 & 68.40\% & 76.36\% & 2.22 & 29.36\% & 42.91\% & 4.62 & 31.50\% & 46.96\% & 4.45 \\
 \addlinespace
 \midrule
 \multirow{3}{*}{\textbf{Ours}}
 & GPT-4.1-mini & 66.30\% & 76.33\% & 2.51 & 31.81\% & 45.97\% & \textbf{\underline{4.25}} & 33.00\% & 45.48\% & \textbf{\underline{4.23}} \\
 & Qwen3-8B & 62.65\% & 74.30\% & 2.90 & 30.56\% & 45.27\% & 4.39 & 32.00\% & 46.78\% & 4.30 \\
 & GPT-5.4 & \underline{71.42\%} & \underline{78.84\%} & \underline{2.20} & \textbf{\underline{33.14\%}} & \textbf{\underline{46.55\%}} & 4.29 & \textbf{\underline{33.50\%}} & 47.44\% & \textbf{\underline{4.23}} \\
 \bottomrule
 \end{tabular}
  
\end{table*}

\paragraph{Baselines.}
We compare \method{} with two supervised sequence models and four LLM-based mobility prediction methods. The supervised baselines include DeepMove~\citep{feng2018deepmove}, an attentional RNN designed to capture periodic mobility patterns and long-term user preferences, and a vanilla Transformer~\citep{vaswani2017attention}. The LLM-based baselines include AgentMove~\citep{feng2025agentmove}, which decomposes zero-shot prediction into spatial-temporal memory, world knowledge, and collective pattern modules; LLM-Mob~\citep{wang2023would}, which formulates prediction as in-context learning over structured trajectory prompts; TrajLLM~\citep{ju2025trajllm}, which converts mobility sequences into textual representations for sequential reasoning; and LLM Urban Residents~\citep{wang2024large}, which models individuals as LLM agents conditioned on activity patterns and retrieved daily motivations. For LLM-based methods, we evaluate Qwen3-8B~\citep{yang2025qwen3}, GPT-4.1-mini~\citep{openai2025gpt41mini}, and GPT-5.4~\citep{openai2026gpt54} as backbones. All methods are evaluated under the same chronological train/test split for each dataset. DeepMove and Transformer are trained only on the training split, while LLM-based baselines and \method{} can access the same training-history records and only the observed context before each target timestamp during inference.

\paragraph{Evaluation Metrics.}
For each test instance, the model outputs a ranked list of candidate locations. We report three metrics: Acc@1, which measures whether the top-ranked prediction matches the ground-truth location; MRR@5, which computes the reciprocal rank of the ground truth within the top five predictions and assigns zero if it is absent; and mean top-1 geographic distance, which measures the Haversine distance in kilometers between the predicted and ground-truth locations.

\subsection{Main Results}
\label{sec:mainresult}

\paragraph{Performance against Baselines.}
Table~\ref{tab:professional_alignment} reports the main comparison on BW, YJMob100K, and the Shanghai ISP first-test-point setting. The Shanghai ISP evaluation follows the same 200-user first-test sample used in prior work~\citep{feng2025agentmove}, and all methods are evaluated under the same chronological split within each dataset. Overall, \method{} achieves the strongest performance among training-free LLM-based methods. With GPT-5.4, \method{} obtains 71.42\% Acc@1, 78.84\% MRR@5, and 2.20\,km distance on BW; 33.14\%, 46.55\%, and 4.29\,km on YJMob100K; and 33.50\%, 47.44\%, and 4.23\,km on Shanghai ISP. These results show that adaptive evidence gathering improves LLM-based mobility prediction across different spatial granularities and observation settings.

Compared with supervised baselines, \method{} does not always dominate task-specific sequence models. On BW, Transformer achieves the best overall Acc@1 and MRR@5, suggesting that supervised training remains highly effective when sufficient regular mobility data are available. However, \method{} with GPT-5.4 outperforms supervised baselines on YJMob100K and Shanghai ISP, where the location space is denser or the available history is shorter. This suggests that evidence-grounded LLM reasoning can be competitive with task-specific training when calibrated mobility statistics, recent context, and location semantics provide useful decision evidence.

Among LLM-based baselines, GPT-5.4 generally improves performance, but the gains vary by method and dataset. LLM Urban Residents remains a strong baseline on BW, likely because its activity-pattern modeling aligns with the stronger temporal regularity of this dataset. AgentMove is competitive on YJMob100K, while \method{} achieves better overall ranking and/or spatial accuracy by explicitly cross-checking behavioral and geographical evidence. The distance metric is particularly informative: on YJMob100K and Shanghai ISP, \method{} avoids the larger spatial drift observed in several prompt-based or scripted LLM baselines. Appendix~\ref{app:baseline_dataset_cases} provides one case for each dataset--baseline pair, showing how these aggregate differences appear at the prediction level.

Shanghai ISP reveals a metric-specific exception. LLM-Mob with GPT-4.1-mini achieves the highest MRR@5, while \method{} achieves higher Acc@1 and lower geographic distance. Since Shanghai ISP contains only eight days of trajectories, several nearby grids can remain plausible for a given user, making top-$K$ ranking easier than selecting the exact top-1 location. \method{} is optimized for the final evidence-grounded decision, which improves top-1 accuracy and spatial error, but does not always yield the most favorable ordering among the remaining plausible candidates. Qualitative trajectory visualizations are provided in Appendix~\ref{app:trajectory_vis}.

\begin{figure}[t]
    \centering
    \includegraphics[width=\columnwidth]{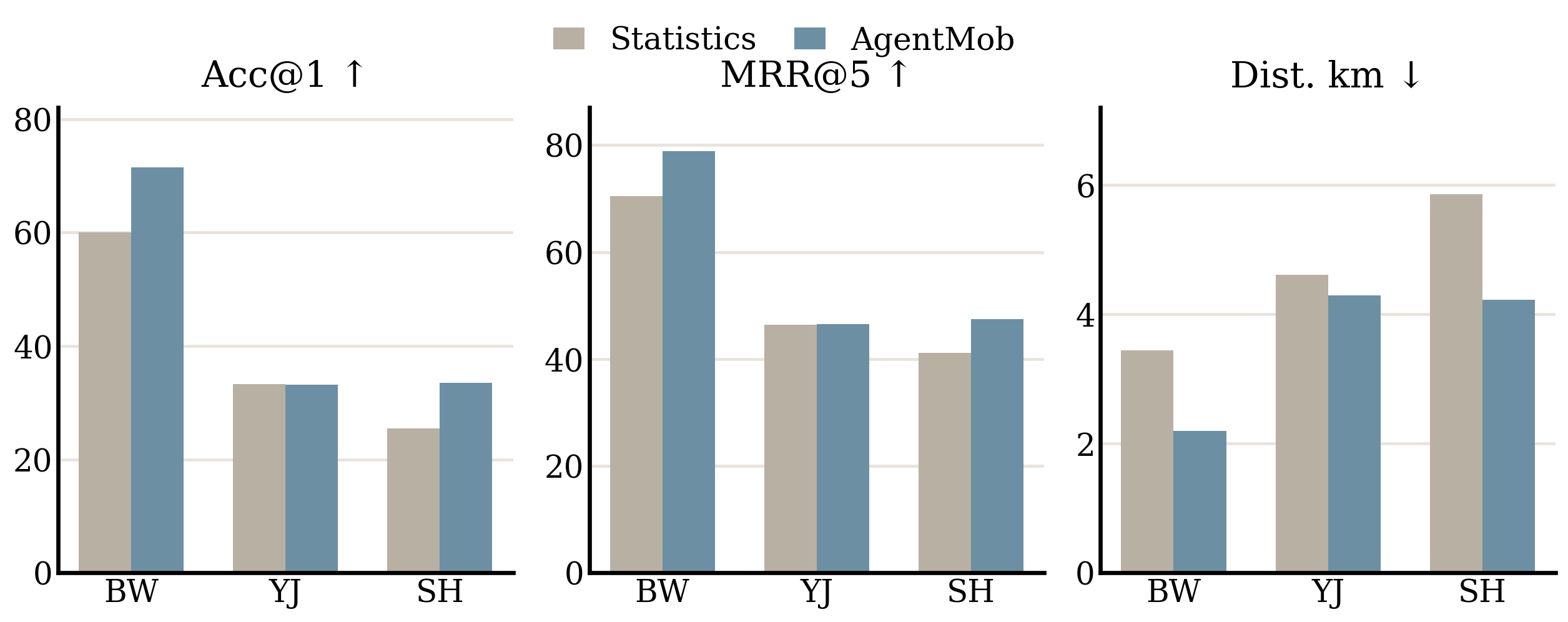}
    \caption{Effect of the LLM controller. \textsc{AgentMob-Statistics} uses the same tool evidence as \method{} but replaces the LLM controller with a deterministic decision rule. Full \method{} uses GPT-5.4.}
    \label{fig:stat_only_baseline}
    \vspace{-1em}
\end{figure}

\paragraph{Reasoning Process Analysis.}
To isolate the contribution of the LLM controller from the underlying mobility statistics, we introduce \textsc{AgentMob-Statistics}, a non-LLM baseline that uses the same tool evidence as \method{} but replaces the LLM controller with a deterministic decision rule. As shown in Figure~\ref{fig:stat_only_baseline}, structured mobility statistics already provide a strong baseline, but the LLM controller further improves decision quality when evidence needs to be reconciled. On BW, \method{} improves Acc@1 from 60.05\% to 71.42\%, MRR@5 from 70.48\% to 78.84\%, and reduces distance from 3.44\,km to 2.20\,km. On Shanghai ISP, Acc@1 increases from 25.50\% to 33.50\%, with distance reduced from 5.86\,km to 4.23\,km.

The gain is smaller on YJMob100K, where \textsc{AgentMob-Statistics} and \method{} are nearly tied in Acc@1. This is likely because YJMob100K uses anonymized grid cells with limited semantic cues, and the structured mobility statistics already capture much of the predictable routine. Nevertheless, \method{} slightly improves MRR@5 and reduces spatial error, suggesting that the controller mainly helps avoid worse off-target predictions rather than changing many exact top-1 decisions.

The benefit of the controller becomes clearer on difficult cases. On the BW non-fast-path subset, where strong historical regularity is absent, \method{} improves Acc@1 from 30.65\% to 48.62\% and MRR@5 from 46.67\% to 60.66\% over \textsc{AgentMob-Statistics}. This supports our central hypothesis: the LLM controller is most useful when deterministic mobility statistics are insufficient and multiple evidence sources must be cross-checked.

Appendix~\ref{app:baseline_dataset_cases} provides representative traces. The BW fast-path case is resolved directly by a perfectly repeated target-hour pattern, while the other cases show \method{} correcting tempting baseline choices by checking hour-specific transitions, local movement evidence, or candidate-ranking signals. These examples support the same mechanism as the aggregate controller results: the agent is most useful when it can cross-check a plausible but weak cue before committing to the final top-1 prediction.

\paragraph{Difficulty-stratified analysis.}


\begin{figure}[t]
    \centering
    \includegraphics[width=\columnwidth]{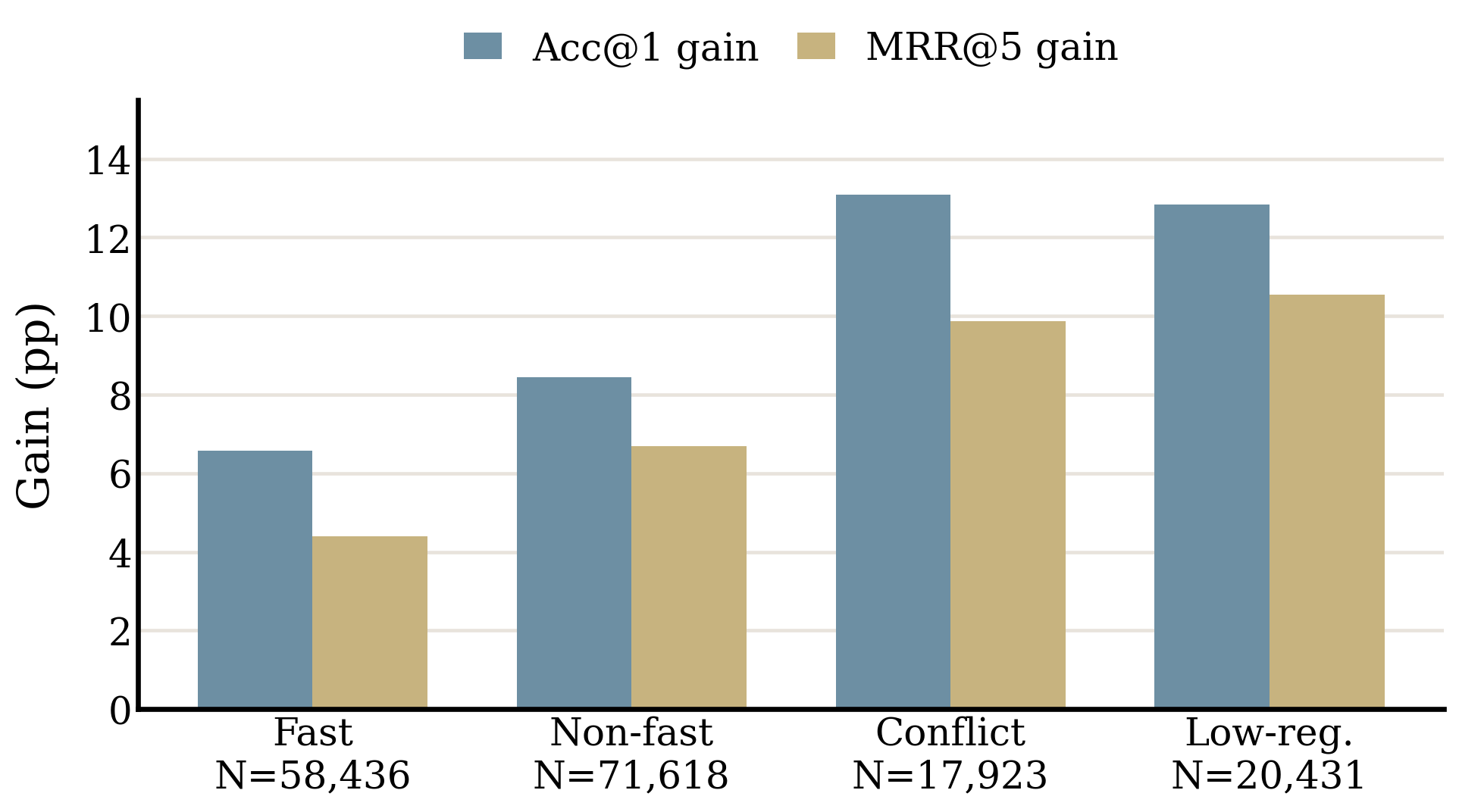}
    \caption{Difficulty-stratified gains of \method{} over \textsc{AgentMob-Statistics} on BW and YJMob100K. The sample size $N$ is shown under each subset.}
    \label{fig:difficulty_analysis}
    \vspace{-1em}
\end{figure}

Figure~\ref{fig:difficulty_analysis} compares \method{} with \textsc{AgentMob-Statistics} across fast-path and harder cases. \method{} improves the accuracy of all cases, but the gains are larger when deterministic evidence is weaker. The least gain of +6.58\% Acc@1 and +4.39\% MRR@5 appears on non-fast-path cases, the most is +13.09\% Acc@1 and +9.87\% MRR@5 under high conflict cases. This indicates the role of the LLM controller in the system. Highly regular cases can often be resolved by structured statistics, whereas difficult cases benefit from the evidence analysis and inference by the LLM controller. When deterministic regularity is weak, candidate scores are close, or historical routines are unreliable, the LLM controller can compare temporal, transition, stay-move, and spatial evidence before making the final prediction. The effect is also dataset-dependent. On YJMob100K and Shanghai ISP, where location information is anonymized grid cell and semantic evidence is limited, the LLM has lesser semantic information beyond the statistics. As a result, the metric improvement of \method{} is not so significant as \textsc{AgentMob-Statistics}.

\paragraph{Efficiency and Cost Analysis.}
Table~\ref{tab:efficiency_trace} reports trace-derived efficiency statistics for GPT-5.4 runs. For each dataset, we compare \method{} with the strongest GPT-5.4 LLM-based baseline in Table~\ref{tab:professional_alignment}. The efficiency behavior reflects the adaptive allocation strategy of \method{}: routine cases can exit through the fast path, while ambiguous cases receive additional tool calls and LLM reasoning.

On BW, 62.54\% of samples are resolved by the fast path, leading to only 1.16 evidence-tool calls per sample and a lower wall-clock time than the strongest LLM baseline. YJMob100K has weaker routine regularity, so more samples enter the tool-calling mode and the average token usage increases. Shanghai ISP has no fast-path exits because each user contributes only one sparse first-test instance, but \method{} still reduces token usage by 94.1\% compared with the strongest GPT-5.4 baseline by replacing long history-heavy prompts with compact structured evidence.

These results show that \method{} is not designed to minimize token usage uniformly across all datasets. Instead, it allocates computation according to prediction difficulty. Even when some non-fast-path samples require more tool reasoning, \method{} achieves lower observed wall-clock time on all three benchmarks under the recorded worker parallelism. Appendix~\ref{app:fast_path_case} provides a fast-path example where a strong target-hour pattern allows the agent to return a prediction without unnecessary tool loops.

\begin{table}[t]
\centering
\scriptsize
\begin{adjustbox}{width=\columnwidth}
\begin{tabular}{llcccc}
\toprule
\textbf{Dataset} & \textbf{Method} & \textbf{Fast path} & \textbf{Tools} & \textbf{Tokens} & \textbf{Wall} \\
 &  &  & /sample & /sample & (s/sample) \\
\midrule
BW & LLM Urban Res. & -- & -- & 4.15k & 2.32 \\
BW & \method{} & 62.54\% & 1.16 & 6.02k & 0.57 \\
\addlinespace
YJMob100K & AgentMove & -- & -- & 1.84k & 1.53 \\
YJMob100K & \method{} & 12.74\% & 2.66 & 12.96k & 0.77 \\
\addlinespace
Shanghai ISP & LLM Urban Res. & -- & -- & 81.40k & 68.77 \\
Shanghai ISP & \method{} & 0.00\% & 1.00 & 4.82k & 2.87 \\
\bottomrule
\end{tabular}
\end{adjustbox}
\caption{Efficiency statistics for GPT-5.4 runs. For each dataset, the baseline is the strongest GPT-5.4 LLM baseline in Table~\ref{tab:professional_alignment}. Tools denote evidence-tool calls in \method{} excluding final answer submission; tokens include prompts and responses; wall-clock time is measured in seconds per sample under the recorded worker parallelism.}
\label{tab:efficiency_trace}
\end{table}


\subsection{Ablations}

\begin{figure}[t]
    \centering
    \includegraphics[width=\columnwidth]{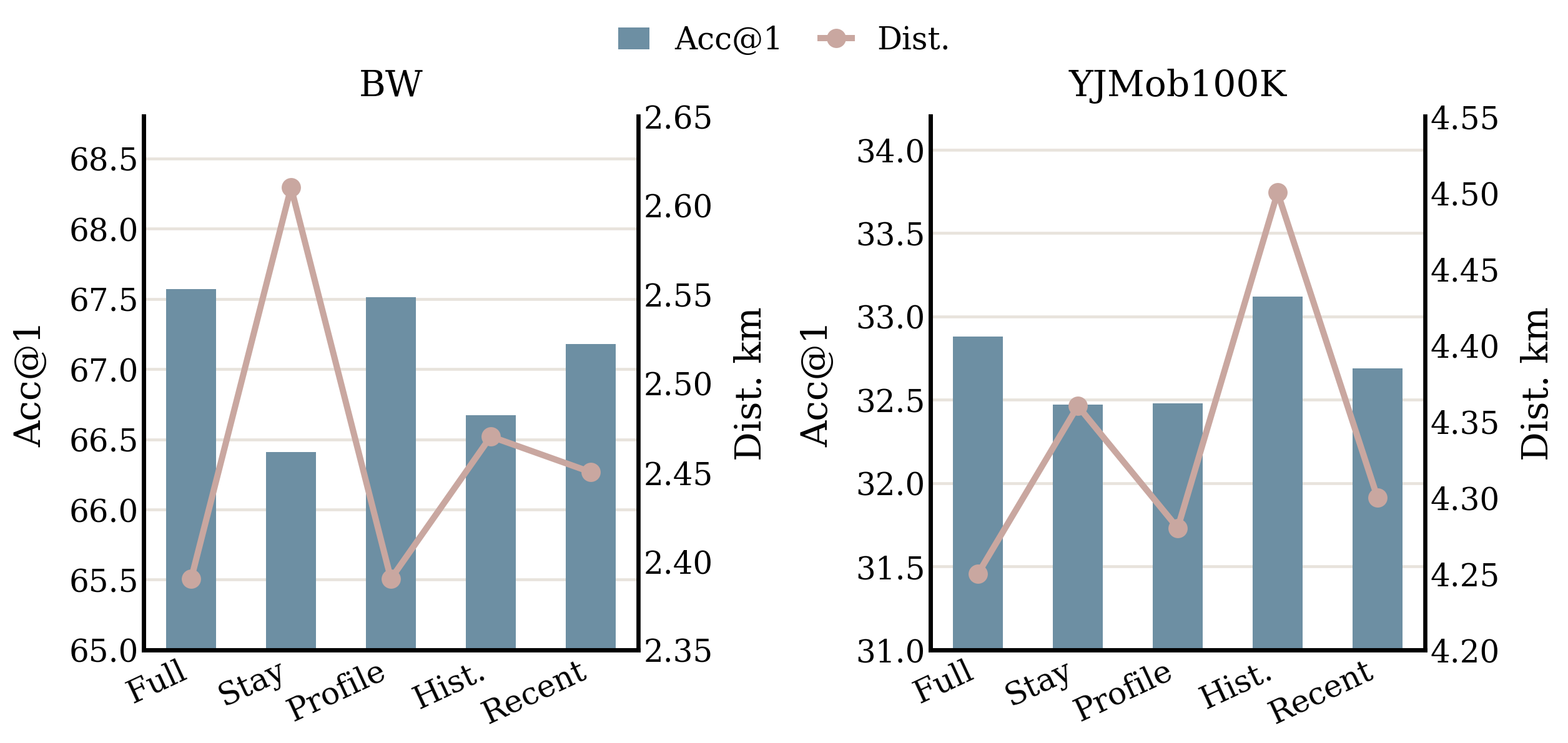}
    \caption{Tool ablation results with GPT-5.4 on BW and YJMob100K, reported before deterministic calibration to isolate the effect of each evidence source. Bars show Acc@1, and lines show geographic distance.}
    \label{fig:ablation_study}
    \vspace{-1em}
\end{figure}

Figure~\ref{fig:ablation_study} reports the effect of removing each evidence source from \method{}. Overall, no single tool dominates all metrics, which is expected because the tools target different types of uncertainty. The full model achieves the most stable overall performance, especially in rank-based accuracy and spatial error. Some ablations can slightly improve Acc@1 on a specific dataset, but they usually worsen MRR@5 or geographic distance, indicating that the removed evidence helps prevent spatially poor alternatives even when the exact top-1 label changes little.

\label{sec:analysis}

\paragraph{Effect of the Stay--Move Estimator.}
Removing the Stay--Move Estimator consistently weakens performance. On BW, Acc@1 drops from 67.57\% to 66.41\%, MRR@5 from 76.90\% to 76.27\%, and distance increases from 2.39\,km to 2.61\,km. On YJMob100K, Acc@1 decreases from 32.88\% to 32.47\%, with distance increasing from 4.25\,km to 4.36\,km. This shows that stay--move evidence is useful for boundary cases where recent continuity and historical departure patterns disagree. A paired trace is provided in Appendix~\ref{app:ablation_cases}.

\paragraph{Effect of the Historical Behavior Retriever.}
The Historical Behavior Retriever provides location-specific evidence such as visit frequency, dwell time, and frequent next destinations. Removing it reduces BW Acc@1 from 67.57\% to 66.67\% and increases distance from 2.39\,km to 2.47\,km. On YJMob100K, Acc@1 slightly increases from 32.88\% to 33.12\%, but distance worsens from 4.25\,km to 4.50\,km. This suggests that historical behavior evidence is especially useful for spatial grounding: even when exact top-1 accuracy changes little, it helps avoid farther off-target predictions in dense grid settings. Additional cases are shown in Appendix~\ref{app:ablation_cases}.

\paragraph{Effect of the Mobility Context Retriever.}
The Mobility Context Retriever supplies recent trajectory context and same-time historical visitation statistics. Removing it causes smaller but consistent degradation: BW Acc@1 decreases from 67.57\% to 67.18\%, and YJMob100K Acc@1 decreases from 32.88\% to 32.69\%, with distance also increasing on both datasets. This indicates that long-term regularity already solves many routine cases, but recent context remains useful when the user has just departed, returned, or stayed unusually long. A paired trace is provided in Appendix~\ref{app:ablation_cases}.

\paragraph{Effect of the Location Profiler.}
The Location Profiler serves as an optional verification tool for candidate locations. Removing it causes only a small drop on BW, from 67.57\% to 67.51\% Acc@1, but a clearer drop on YJMob100K, from 32.88\% to 32.48\%. This matches its role in ambiguous cases: it helps check whether a candidate is spatially plausible and semantically consistent before the agent changes its decision. A paired trace is provided in Appendix~\ref{app:ablation_cases}.

\paragraph{Sensitivity to the backbone LLM.}
\method{} also depends on the backbone model's tool-use ability. Compared with GPT-4.1-mini, Qwen3-8B shows weaker multi-tool coordination: after entering tool-calling mode, 29.1\% of cases fail to invoke the Stay--Move Estimator as prescribed. Since reliable tool use requires instruction following, API selection, and multi-step planning~\citep{qin2023toolllm}, smaller or less aligned models may require stricter tool-call validation, simplified orchestration, or distillation of tool-use behavior. This is consistent with reported model-scale variation in Qwen3 instruction following~\citep{yang2025qwen3}.


\section{Conclusion and Future Work}
\label{sec:conclusion}

We presented \method{}, a training-free LLM-agent that formulates next-location prediction as adaptive evidence-controlled decision making. Instead of relying on static prompt-only inference, \method{} allocates reasoning effort according to prediction difficulty. Routine cases are resolved through a fast path, while ambiguous cases trigger selective tool use over recent context, historical behavior, stay--move likelihood, and geographical evidence. Experiments on different datasets show \method{} achieves strong training-free performance while producing auditable prediction traces. Further analyses demonstrate that the LLM controller is most useful when deterministic mobility statistics are insufficient or conflicting. Future work includes stronger uncertainty estimation, risk-aware mobility planning, and broader multi-city transfer.

\section{Limitations}

We discuss several scope boundaries of the current study and how they motivate future extensions.

First, \method{} relies on the backbone LLM's ability to follow tool-use instructions and coordinate multi-step reasoning. This is inherent to tool-augmented agent frameworks rather than specific to mobility prediction. Our backbone analysis shows that stronger instruction-following models can better exploit the evidence-control protocol, while smaller models may require additional safeguards. Future work could improve robustness through stricter tool-call validation, simplified orchestration, or distillation of successful tool-use traces into smaller models.

Second, our evaluation focuses on automatically collected mobility trajectories, including continuous GPS datasets and a sparse mobile-network benchmark. This choice matches our goal of studying next-location prediction under chronological constraints from passively sensed trajectories. We therefore do not directly evaluate on social check-in datasets such as Foursquare, where locations are actively reported events and follow a different observation mechanism from continuous sensing. Extending \method{} to bridge check-in data, continuous GPS traces, and mobile-network records is an important direction for broader generalization.

Finally, the current toolbox is designed around general evidence categories for mobility prediction: recent context, historical behavior, stay--move likelihood, and geographical plausibility. Although these tools are effective across the studied datasets, their boundaries and invocation policy are manually specified. More systematic optimization of tool design and orchestration could further improve performance, for example through automated prompt tuning, learned tool-selection policies, or dynamic tool composition. These extensions are complementary to our core contribution of formulating mobility prediction as adaptive evidence-controlled decision making.

\section{Risk Statement}
The current tool design is based on the capabilities of existing LLMs and may become less effective as LLM capabilities evolve.

\bibliography{reference}

\appendix
\section{Qualitative Trajectory Visualization}
\label{app:trajectory_vis}
\begin{figure}[h]
    \centering
    \includegraphics[width=\columnwidth]{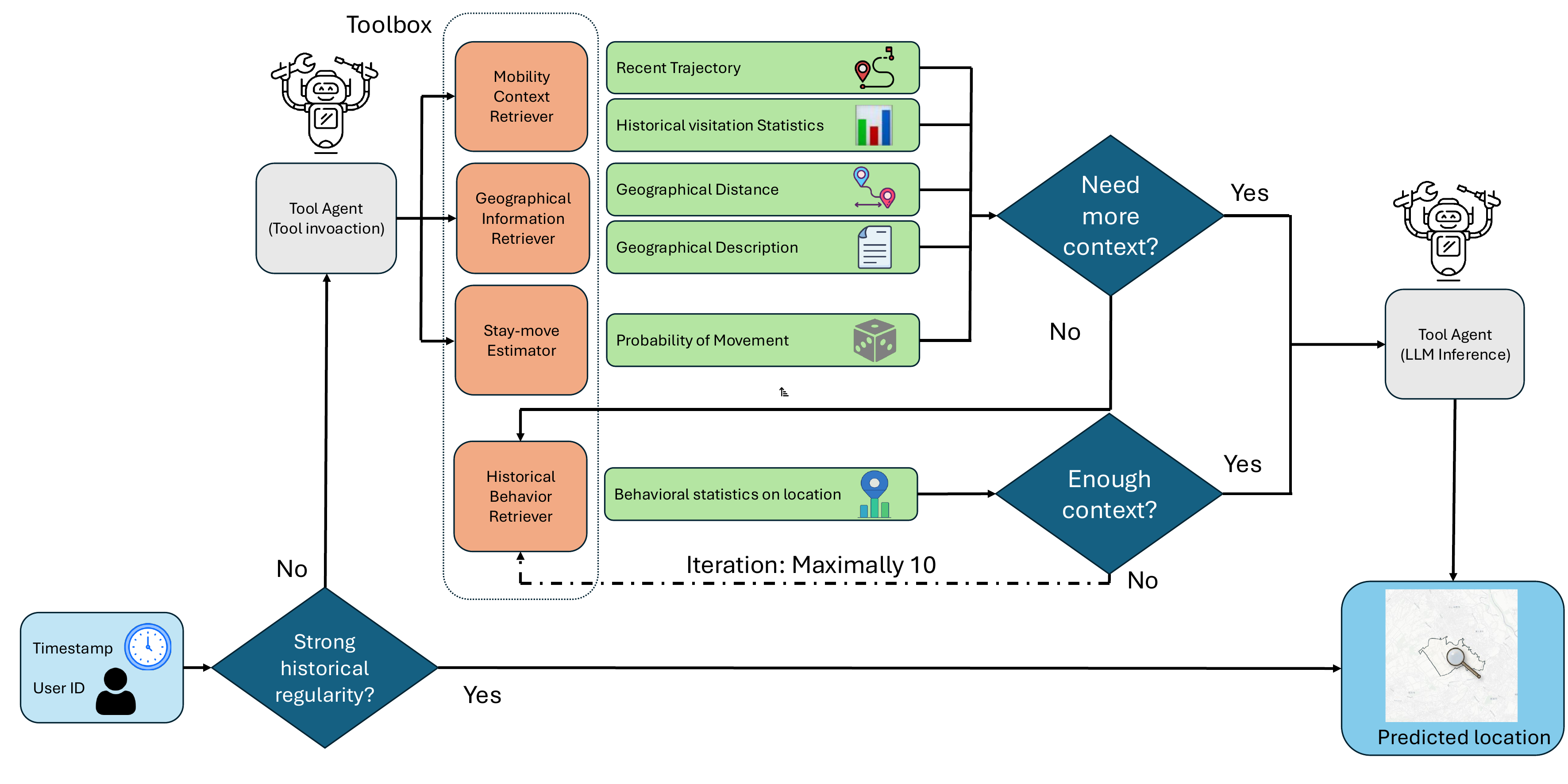}
    \caption{Part of predictions for a sample user. (a) Ground truth (b) Our method (c-e) Baselines.}
    \label{fig:trajectory_case}
\end{figure}

\section{Location Description prediction}
\label{app:loc_desc}

Since raw GPS coordinates and anonymous cell IDs carry no semantic information, we design a description-prediction step that converts each spatial unit into a concise textual characterization, enabling the LLM to reason about locations based on their functional roles.
All descriptions are generated once with GPT-4o and shared across every LLM-based method evaluated in this paper.

\paragraph{BW.}
Each polygon corresponds to a third-level administrative unit and is identified by a hierarchical place name (prefecture / municipality / district, e.g., \textit{T\=oky\=o-to, Itabashi-ku, Hasune}).
We retrieve publicly available geographic information for each place name and prompt GPT-4o to produce a one-sentence functional summary.
For example, the polygon \textit{T\=oky\=o-to, Itabashi-ku, Hasune} yields:

\begin{quote}\small
\textit{``It functions primarily as a quiet residential neighborhood, where the area is characterized by apartment complexes, small parks, and local convenience stores, with nearby transit access via Hasune Station on the Mita Line.''}
\end{quote}

\paragraph{YJMob100K.}
Cells are anonymized 500\,m\,$\times$\,500\,m grid squares with no place names.
We leverage the POI category vectors (85 categories) provided by the dataset authors~\citep{yabe2024yjmob100k}, and prompt GPT-4o to summarize each cell's top POI categories and their relative proportions into a one-sentence description.
For example, a cell whose top categories are \textit{Hair Salon} (8.3\%), \textit{Transit Station} (7.4\%), and \textit{Hospital} (7.0\%) yields:

\begin{quote}\small
\textit{``It functions primarily as a diverse mixed-use service area, where the mapped profile is led by many hair salons alongside many transit stations and many hospitals, with adjacent cells indicating a more active, service-rich amenity landscape and no single activity clearly dominating the amenity mix.''}
\end{quote}

\section{Ablation Case Examples}
\label{app:ablation_cases}

\paragraph{Stay-move Estimator.}
For user \texttt{fef58e6c6288Bad60f5629111c489116} at 2023-06-16 07:00 in BW, the no-stay-move ablation predicts \textcolor{red}{76620}, while the full method predicts the ground-truth location \textcolor{red}{17833}. The paired traces show why the Stay-move Estimator matters.

\noindent\textbf{No-stay-move ablation trace}
\begin{quote}
\textit{``The user has stayed at 76620 for the last five hours, transition continuation from the current location strongly favors remaining at 76620 (88.4\%), and the top composite score is also 76620. Although the Friday 07:00 historical signal points to 17833, it is based on only one prior instance and is weaker than the strong current stay/transition evidence.''}
\end{quote}

\noindent\textbf{Full-method trace}
\begin{quote}
\textit{``Average stay: 8.6h. Current stay: 5h (0.58x average). Typical departure hours: 7:00. Current hour 7:00 matches a typical departure hour. Transitions at 07:00 from Location 76620: 17833: 64\%; 76620 (stay): 24\%.''}
\end{quote}

\begin{quote}
\textit{``Although 76620 is the top raw candidate, multiple stronger move signals point to 17833: 07:00 is a typical departure hour from 76620, hour-specific transitions from 76620 go to 17833 64\% vs only 24\% stay, and the only past Friday 07:00 example is also 17833. These aligned commute-time signals justify overriding the rank-1 stay candidate.''}
\end{quote}

This case supports the aggregate ablation result: without the Stay-move Estimator, the agent overweights recent continuity; with the estimator, it can identify departure-boundary cases where the user is likely to move despite several recent hours at the same location.

\paragraph{Mobility Context Retriever.}
For user \texttt{ed1fedcdf0c95e6748f68ee552aba4d2} at 2023-05-25 07:00 in BW, the no-recent-trajectory ablation predicts \textcolor{red}{35995}, while the full method predicts the ground-truth location \textcolor{red}{60203}. The paired traces show that recent context prevents the agent from overreacting to a sparse weekday-hour pattern.

\noindent\textbf{No-recent-trajectory ablation trace}
\begin{quote}
\textit{``Although the current-location stay signal is strong, the strongest Tier-1 evidence is the past same Thursday 07:00 example and the weekly 07:00 pattern, both pointing to 35995. Because this is a commute window and the dominant day+hour location differs from the current location, I override the rank-1 candidate and predict 35995.''}
\end{quote}

\noindent\textbf{Full-method trace}
\begin{quote}
\textit{``Recent Trajectory (last 6): 2023-05-25T01:00:00: ID 60203 ... 2023-05-25T06:00:00: ID 60203. Last 4 hours: ID 60203 $\rightarrow$ 60203 $\rightarrow$ 60203 $\rightarrow$ 60203 (stable).''}
\end{quote}

\begin{quote}
\textit{``Although past Thursday 07:00 examples favor 35995, the user has been settled at 60203 for 7 hours, current stay matches the historical average, and both overall and 07:00-specific transitions strongly favor remaining at 60203. Given the stable overnight trajectory and 78\% self-transition at 07:00, staying at 60203 is the most likely outcome.''}
\end{quote}

This case supports the ablation result: without recent trajectory, the agent follows the same weekday-hour memory; with the Mobility Context Retriever, it sees that the user is in a stable local continuation state.

\paragraph{Historical Behavior Retriever.}
For user \texttt{b1774dafe8Ba1caac3e7aeb1b2b001d4} at 2023-06-21 13:00 in BW, the no-retriever ablation predicts \textcolor{red}{33524}, while the full method predicts the ground-truth location \textcolor{red}{96817}. The paired traces show that the Historical Behavior Retriever helps the agent audit whether a sparse historical signal corresponds to a truly visited location.

\noindent\textbf{No-retriever ablation trace}
\begin{quote}
\textit{``The strongest signal is the Wednesday 13:00 historical pattern, where ID 33524 accounts for 50.0\% of past instances and also receives the strongest initial support. Current-location stay evidence is weak with no transition or duration history for 96817, so the dominant same-day/hour pattern should override staying nearby.''}
\end{quote}

\noindent\textbf{Full-method trace}
\begin{quote}
\textit{``Location Details: 96818. Visit frequency: 0.00\% (0 visits).''}
\end{quote}

\begin{quote}
\textit{``The strong Wednesday 13:00 historical signal points to 96818, but that location is not in the candidate list and shows no visit history in the details lookup. With no transition data from 96817 and the user only having stayed there 1 hour after local movement, I choose the only plausible nearby visited continuation among candidates: remaining at current location 96817, though with low confidence.''}
\end{quote}

This case supports the ablation result: without the retriever, the agent follows a coarse same-day/hour pattern; with the retriever, it can check the visitation evidence behind that pattern and avoid trusting an unsupported historical candidate.

\paragraph{Location Profiler.}
For user \texttt{5c6fe37f6005d3aeb4e3062f44bb32d3} at 2023-06-01 12:00 in BW, the no-optional-tools ablation predicts \textcolor{red}{55472}, while the full method predicts the ground-truth location \textcolor{red}{37601}. The paired traces show that the Location Profiler helps the agent reject an unsupported historical override before leaving a stable current location.

\noindent\textbf{No-optional-tools ablation trace}
\begin{quote}
\textit{``The user has been stationary at 37601 for 24 hours, but this far exceeds the historical average stay and matches typical departure timing, so move signals are strong. Among plausible move destinations, 55472 has the highest composite score and is strongly supported by hourly and overall visitation patterns, while the Thursday-12:00 historical signal for 66619 is based on only one instance and is not in the candidate set.''}
\end{quote}

\noindent\textbf{Full-method trace}
\begin{quote}
\textit{``Location Details: 66619. Visit frequency: 0.00\% (0 visits).''}
\end{quote}

\begin{quote}
\textit{``Despite the top composite score for 55472, the user has remained at 37601 for 24 hours with a settled pattern and all recent hours unchanged. The strong Thursday-12:00 signal points to 66619, but that location has 0 recorded visits and is not a valid candidate, so continuation/self-transition to 37601 is the safest prediction.''}
\end{quote}

This case supports the ablation result: optional profiling is not needed for every sample, but it helps in ambiguous cases by checking whether a tempting alternative is supported by concrete location-level evidence.

\section{Inference Case Studies}
\label{app:inference_cases}

\subsection{Baseline-by-Dataset Comparison Cases}
\label{app:baseline_dataset_cases}

This section gives one representative comparison for each dataset-baseline pair. Each case reports the ground truth, \method{}'s prediction, and the baseline's top-ranked prediction, followed by a short explanation of the decision difference.

\paragraph{BW--AgentMove.}\mbox{}\\
\label{app:fast_path_case}
\noindent\textbf{Dataset:} BW\\
\noindent\textbf{Ground truth:} \textcolor{red}{36166} \\
\textbf{Ours:} {correct} \\
\textbf{AgentMove:} wrong (\textcolor{red}{17844} ranked first)

\vspace{0.5em}
\noindent\textbf{Ours inference}
\begin{quote}\small
The prediction is made directly through the \textcolor{red}{fast-path mechanism}. The temporal evidence shows a \textcolor{red}{perfectly repeated pattern}: on Friday at 09:00, the user was at location \textcolor{red}{36166} in \textcolor{red}{18 out of 18} historical instances. Since the \textcolor{red}{day-hour pattern is fully consistent}, the model can confidently rely on this structured temporal signal without invoking additional tools.
\end{quote}

\noindent\textbf{AgentMove inference}
\begin{quote}\small
The user's most visited location is \textcolor{red}{17844} with a high overall visit rate and a significant presence in their long-term history, indicating a strong likelihood to return especially during weekday mornings. \\
... \textcolor{red}{36166} ... is ranked third.
\end{quote}

\paragraph{BW--LLM-Mob.}\mbox{}\\
\noindent\textbf{Dataset:} BW\\
\noindent\textbf{Ground truth:} \textcolor{red}{84174} \\
\textbf{Ours:} {correct} \\
\textbf{LLM-Mob:} wrong (\textcolor{red}{60263} ranked first)

\vspace{0.5em}
\noindent\textbf{Ours inference}
\begin{quote}\small
The candidate ranking step corrected the initial prediction and selected \textcolor{red}{84174}. Although several overnight locations were plausible, the final evidence-supported candidate was the ground-truth location.
\end{quote}

\noindent\textbf{LLM-Mob inference}
\begin{quote}\small
LLM-Mob emphasized the Wednesday 01:00 historical pattern and the immediately preceding trajectory, ranking \textcolor{red}{60263} first because it appeared to match overnight behavior. It treated \textcolor{red}{84174} as a later early-morning location and ranked it below the top candidates.
\end{quote}

\paragraph{BW--LLM Urban Residents.}\mbox{}\\
\noindent\textbf{Dataset:} BW\\
\noindent\textbf{Ground truth:} \textcolor{red}{4419} \\
\textbf{Ours:} {correct} \\
\textbf{LLM Urban Residents:} wrong (\textcolor{red}{76623} ranked first)

\vspace{0.5em}
\noindent\textbf{Ours inference}
\begin{quote}\small
\method{} used the hour-specific transition evidence to move away from the initial candidate and selected \textcolor{red}{4419}. The decision follows the local early-morning evidence rather than the broader persona-level routine.
\end{quote}

\noindent\textbf{LLM Urban Residents inference}
\begin{quote}\small
LLM Urban Residents emphasized a weekend overnight movement pattern and the user's broader activity persona, ranking \textcolor{red}{76623} first as a likely early-morning return location. The ground-truth location \textcolor{red}{4419} was kept as a lower-ranked recent stop.
\end{quote}

\paragraph{YJMob100K--AgentMove.}\mbox{}\\
\noindent\textbf{Dataset:} YJMob100K\\
\noindent\textbf{Ground truth:} \textcolor{red}{(134,83)} \\
\textbf{Ours:} {correct} \\
\textbf{AgentMove:} wrong (\textcolor{red}{(155,106)} ranked first)

\vspace{0.5em}
\noindent\textbf{Ours inference}
\begin{quote}\small
The candidate ranking step selected \textcolor{red}{(134,83)} after comparing the immediate candidate evidence. This prevents the prediction from collapsing to a distant long-term anchor when the current sample favors a different location.
\end{quote}

\noindent\textbf{AgentMove inference}
\begin{quote}\small
AgentMove prioritized the user's dominant long-term anchor \textcolor{red}{(155,106)}, which accounts for a large share of visits and has strong self-transition behavior. This broad routine signal outweighed the sample-specific evidence for \textcolor{red}{(134,83)}.
\end{quote}

\paragraph{YJMob100K--LLM-Mob.}\mbox{}\\
\noindent\textbf{Dataset:} YJMob100K\\
\noindent\textbf{Ground truth:} \textcolor{red}{(107,75)} \\
\textbf{Ours:} {correct} \\
\textbf{LLM-Mob:} wrong (\textcolor{red}{(98,78)} ranked first)

\vspace{0.5em}
\noindent\textbf{Ours inference}
\begin{quote}\small
\method{} corrected the initial nearby candidate and selected \textcolor{red}{(107,75)}. The final decision stays within the current local movement corridor rather than following a repeated afternoon pattern from another grid cluster.
\end{quote}

\noindent\textbf{LLM-Mob inference}
\begin{quote}\small
LLM-Mob emphasized a repeated historical pattern around \textcolor{red}{(98,78)}, ranking that location first. It considered the \textcolor{red}{107}-cluster plausible but placed it below the stronger routine signal.
\end{quote}

\paragraph{YJMob100K--LLM Urban Residents.}\mbox{}\\
\noindent\textbf{Dataset:} YJMob100K\\
\noindent\textbf{Ground truth:} \textcolor{red}{(169,126)} \\
\textbf{Ours:} {correct} \\
\textbf{LLM Urban Residents:} wrong (\textcolor{red}{(138,89)} ranked first)

\vspace{0.5em}
\noindent\textbf{Ours inference}
\begin{quote}\small
The candidate ranking step selected \textcolor{red}{(169,126)}, preserving the evidence for the current work-related location rather than extrapolating from the latest movement cluster.
\end{quote}

\noindent\textbf{LLM Urban Residents inference}
\begin{quote}\small
LLM Urban Residents relied on the recent movement within the \textcolor{red}{(138,88)--(139,88)} cluster and a factory-worker persona, ranking \textcolor{red}{(138,89)} first as a likely continuation. The ground-truth location was included only as a lower-ranked alternative.
\end{quote}

\paragraph{Shanghai ISP--AgentMove.}\mbox{}\\
\noindent\textbf{Dataset:} Shanghai ISP\\
\noindent\textbf{Ground truth:} \textcolor{red}{(138,114)} \\
\textbf{Ours:} {correct} \\
\textbf{AgentMove:} wrong (\textcolor{red}{(129,121)} ranked first)

\vspace{0.5em}
\noindent\textbf{Ours inference}
\begin{quote}\small
\method{} kept the base prediction \textcolor{red}{(138,114)}, indicating that the available evidence did not justify moving to a farther historical anchor in this sparse first-test setting.
\end{quote}

\noindent\textbf{AgentMove inference}
\begin{quote}\small
AgentMove emphasized the user's overall preference for \textcolor{red}{(129,121)} and frequent transitions around that area. This long-term anchor dominated its prediction, even though the target remained at \textcolor{red}{(138,114)}.
\end{quote}

\paragraph{Shanghai ISP--LLM-Mob.}\mbox{}\\
\noindent\textbf{Dataset:} Shanghai ISP\\
\noindent\textbf{Ground truth:} \textcolor{red}{(92,105)} \\
\textbf{Ours:} {correct} \\
\textbf{LLM-Mob:} wrong (\textcolor{red}{(71,91)} ranked first)

\vspace{0.5em}
\noindent\textbf{Ours inference}
\begin{quote}\small
\method{} used hour-transition calibration to revise the base prediction from \textcolor{red}{(71,91)} to \textcolor{red}{(92,105)}. The final answer favors the target-time evidence over a repeated but less reliable recent morning pattern.
\end{quote}

\noindent\textbf{LLM-Mob inference}
\begin{quote}\small
LLM-Mob ranked \textcolor{red}{(71,91)} first because the user appeared there at 08:00 on two recent days. It still ranked \textcolor{red}{(92,105)} second, but did not elevate it to the final top-1 prediction.
\end{quote}

\paragraph{Shanghai ISP--LLM Urban Residents.}\mbox{}\\
\noindent\textbf{Dataset:} Shanghai ISP\\
\noindent\textbf{Ground truth:} \textcolor{red}{(125,108)} \\
\textbf{Ours:} {correct} \\
\textbf{LLM Urban Residents:} wrong (\textcolor{red}{(126,107)} ranked first)

\vspace{0.5em}
\noindent\textbf{Ours inference}
\begin{quote}\small
\method{} used hour-transition calibration to select \textcolor{red}{(125,108)} instead of the neighboring morning anchor \textcolor{red}{(126,107)}. The correction is small spatially but changes the exact top-1 label.
\end{quote}

\noindent\textbf{LLM Urban Residents inference}
\begin{quote}\small
LLM Urban Residents ranked \textcolor{red}{(126,107)} first because it was a regular 08:00 location across recent similar days. It placed \textcolor{red}{(125,108)} second, showing that the baseline identified the right local area but missed the exact grid.
\end{quote}

\end{document}